\documentclass[conference]{IEEEtran}
\IEEEoverridecommandlockouts

\usepackage{cite}
\usepackage{amsmath,amssymb,amsfonts}
\usepackage{algorithmic}
\usepackage{graphicx}
\usepackage{textcomp}
\usepackage{xcolor}
\def\BibTeX{{\rm B\kern-.05em{\sc i\kern-.025em b}\kern-.08em
    T\kern-.1667em\lower.7ex\hbox{E}\kern-.125emX}}
\begin{document}

\title{Predictive Maintenance of Armoured Vehicles using Machine Learning Approaches\\ 
{\footnotesize \textsuperscript{}}
\thanks{*Corresponding Author: amehta1\_be20@thapar.edu}
}

\author{\IEEEauthorblockN{ Prajit Sengupta}
\IEEEauthorblockA{\textit{CSED} \\
\textit{Thapar Institute of}\\
\textit{Engineering and Technology}\\
Patiala, India \\
psengupta\_be20@thapar.edu}
\and
\IEEEauthorblockN{Anant Mehta*}
\IEEEauthorblockA{\textit{CSED} \\
\textit{Thapar Institute of}\\
\textit{Engineering and Technology}\\
Patiala, India \\
amehta1\_be20@thapar.edu}
\and
\IEEEauthorblockN{Prashant Singh Rana}
\IEEEauthorblockA{\textit{CSED} \\
\textit{Thapar Institute of}\\
\textit{Engineering and Technology}\\
Patiala, India \\
prashant.singh@thapar.edu}

}

\maketitle

\begin{abstract}
Armoured vehicles are specialized and complex pieces of machinery designed to operate in high-stress environments, often in combat or tactical situations.
This study proposes a predictive maintenance-based ensemble system that aids in predicting potential maintenance needs based on sensor data collected from these vehicles. The proposed model's architecture involves various models such as Light Gradient Boosting, Random Forest, Decision Tree, Extra Tree Classifier and Gradient Boosting to predict the maintenance requirements of the vehicles accurately. In addition, K-fold cross validation, along with TOPSIS analysis, is employed to evaluate the proposed ensemble model's stability. The results indicate that the proposed system achieves an accuracy of 98.93\%, precision of 99.80\% and recall of 99.03\%. The algorithm can effectively predict maintenance needs, thereby reducing vehicle downtime and improving operational efficiency. Through comparisons between various algorithms and the suggested ensemble, this study highlights the potential of machine learning-based predictive maintenance solutions.
\end{abstract}


\section{Introduction}
Armoured vehicles are critical assets in defence and security operations, and their proper functioning is essential for mission success. These vehicles, which are designed to withstand hostile environments and provide protection for personnel and equipment, are complex systems that rely on a wide range of components, including engines, transmissions, and hydraulics. These components can experience wear and tear over time, leading to potential failures and downtime that can compromise the effectiveness of the vehicle\cite{borkowski2010influence}. Traditional maintenance approaches often rely on scheduled maintenance and inspections, which can be inefficient and costly. Predictive maintenance, on the other hand, can leverage machine learning algorithms to identify potential problems and optimize maintenance schedules, reducing costs and downtime while improving the overall reliability and effectiveness of the vehicle. \\ \\
Machine learning approaches, such as supervised and unsupervised learning, by analyzing massive volumes of data and finding patterns that might reveal future faults, can be useful in predictive maintenance. 
\begin{figure}[ht]
\centerline{\includegraphics[scale=0.7]{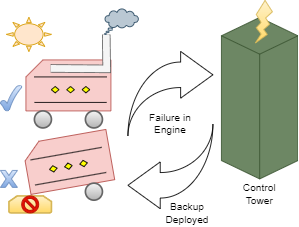}}
\caption{Communication between broken-down vehicle and control tower}
\label{fig_intro}
\end{figure}
This approach can reduce maintenance costs, improve vehicle readiness, and enhance overall operational efficiency. This study explores the potential of ensemble model based approach for predictive maintenance of armoured vehicles\cite{dong2020survey}. A meta algorithm is proposed that is comprised of five base classifiers, to help in technical surveillance as shown in figure \ref{fig_intro}. A thorough examination of the pertinent literature is conducted, along with a discussion of the relative merits and demerits of the various existing machine learning algorithms used for armoured vehicle predictive maintenance. This research also discusses the challenges and future prospects of this emerging field, highlighting areas where further research can be done. By doing so, this paper provides a valuable resource for researchers, practitioners, and policymakers interested in improving armoured vehicles' maintenance and operational readiness.

\begin{table*}[!ht]
\centering
\caption{Feature Table}
\begin{tabular}{|c|l|l|l|l|l|l|c|}
\hline
\textbf{Machine Number} & \textbf{F1} & \textbf{F2 [K]}          & \textbf{F3 [K]}         & \textbf{F4 [rpm]}         & \textbf{F5 [Nm]}       & \textbf{F6 [min]}          & \textbf{Machine failure}  \\
\hline
0              & 2  & 0.304347826 & 0.358024691 & 0.222933644 & 0.535714286 & 0           & 0               \\
1              & 1  & 0.315217391 & 0.37037037  & 0.139697322 & 0.583791209 & 0.011857708 & 0               \\
2              & 1  & 0.304347826 & 0.345679012 & 0.192083818 & 0.626373626 & 0.019762846 & 0               \\
3              & 1  & 0.315217391 & 0.358024691 & 0.154249127 & 0.490384615 & 0.027667984 & 0               \\
4              & 1  & 0.315217391 & 0.37037037  & 0.139697322 & 0.497252747 & 0.035573123 & 0               \\
5              & 2  & 0.304347826 & 0.358024691 & 0.149592549 & 0.523351648 & 0.043478261 & 0               \\
6              & 1  & 0.304347826 & 0.358024691 & 0.227008149 & 0.53021978  & 0.055335968 & 0               \\
7              & 1  & 0.304347826 & 0.358024691 & 0.208963912 & 0.5         & 0.063241107 & 0               \\
8              & 2  & 0.326086957 & 0.37037037  & 0.290454016 & 0.340659341 & 0.071146245 & 0               \\
9              & 2  & 0.347826087 & 0.407407407 & 0.333527357 & 0.332417582 & 0.083003953 & 0               \\
.              &    & .           &             & .           &             & .           &                 \\
.              &    & .           &             & .           &             & .           &                 \\
1353           & 2  & 0.358695652 & 0.543209877 & 0.160651921 & 0.608516484 & 0.154150198 & 0               \\
1354           & 1  & 0.358695652 & 0.530864198 & 0.096623981 & 0.607142857 & 0.166007905 & 0               \\
1355           & 2  & 0.358695652 & 0.530864198 & 0.298603027 & 0.43543956  & 0.173913043 & 0               \\
1356           & 1  & 0.358695652 & 0.518518519 & 0.247380675 & 0.461538462 & 0.185770751 & 0               \\
1357           & 1  & 0.358695652 & 0.518518519 & 0.170547148 & 0.575549451 & 0.193675889 & 0               \\
1358           & 2  & 0.369565217 & 0.543209877 & 0.305587893 & 0.385989011 & 0.201581028 & 0               \\
1359           & 1  & 0.369565217 & 0.555555556 & 0.137369034 & 0.638736264 & 0.213438735 & 0               \\
1360           & 0  & 0.380434783 & 0.555555556 & 0.313736903 & 0.326923077 & 0.221343874 & 0               \\
.              &    & .           &             & .           &             & .           &                 \\
.              &    & .           &             & .           &             & .           &                 \\
7196           & 0  & 0.52173913  & 0.543209877 & 0.150174622 & 0.534340659 & 0.252964427 & 0               \\
7197           & 1  & 0.510869565 & 0.530864198 & 0.185098952 & 0.521978022 & 0.272727273 & 0               \\
7198           & 2  & 0.510869565 & 0.518518519 & 0.210710128 & 0.421703297 & 0.280632411 & 0               \\
7199           & 1  & 0.510869565 & 0.518518519 & 0.197322468 & 0.516483516 & 0.292490119 & 0               \\
7200           & 1  & 0.510869565 & 0.518518519 & 0.256111758 & 0.406593407 & 0.300395257 & 0               \\
7201           & 1  & 0.510869565 & 0.50617284  & 0.178114086 & 0.505494505 & 0.308300395 & 0               \\
7202           & 1  & 0.510869565 & 0.49382716  & 0.091967404 & 0.798076923 & 0.316205534 & 0               \\
.              &    & .           &             & .           &             & .           &                 \\
.              &    & .           &             & .           &             & .           &                 \\
9992           & 1  & 0.380434783 & 0.333333333 & 0.183934808 & 0.486263736 & 0.031620553 & 0               \\
9993           & 1  & 0.380434783 & 0.333333333 & 0.135622817 & 0.597527473 & 0.039525692 & 0               \\
9994           & 1  & 0.380434783 & 0.320987654 & 0.271245634 & 0.331043956 & 0.04743083  & 0               \\
9995           & 2  & 0.380434783 & 0.333333333 & 0.253783469 & 0.353021978 & 0.055335968 & 0               \\
9996           & 0  & 0.391304348 & 0.333333333 & 0.27008149  & 0.384615385 & 0.067193676 & 0               \\
9997           & 2  & 0.402173913 & 0.358024691 & 0.277648428 & 0.406593407 & 0.086956522 & 0               \\
9998           & 0  & 0.402173913 & 0.37037037  & 0.139697322 & 0.614010989 & 0.098814229 & 0               \\
9999           & 2  & 0.402173913 & 0.37037037  & 0.193247963 & 0.5         & 0.118577075 & 0   
\\
\hline
\end{tabular}
\label{tab1}

\end{table*}

\section{Related Work}
Xiao et al. (2021) explained the basics of machine learning and fault diagnosis in their paper. Several other popular machine learning techniques were discussed, and the development status in recent years was summarised and analyzed\cite{xiao2021review}.
In another study, Theissler et al. (2021) discussed the PdM(Predictive Maintainence), when integrated with Machine Learning approaches provides valuable result. In this study they collected, classified, and analyzed papers from an application and Machine Learning standpoint\cite{theissler2021predictive}.
Arena et al. (2021) provided a comprehensive research study of AI and statistical inference techniques, as well as stochastic methods, for use in automotive preventative maintenance. The authors presented an LSTM network for RUL prediction, which was given features extracted from the sensor data by a CNN (Convolutional Neural Network). \\
In addition to predicting the RUL, machine learning approaches were also used to optimize the maintenance schedules of armoured vehicles\cite{arena2022predictive}. In order to predict a heavy machine's condition for maintenance, this paper presented by Putra et al. (2021) concentrated on developing ML models with actual data from large vehicles like trains and tanks. Classification Algorithms were used to predict the failure scope of the machine in order to maintain it\cite{putra2021designing}.
A study by Jain et al. (2022) used a Deep Learning method to estimate how long a tank engine would last in service\cite{jain2022systematic}. Similarly, a study by Tessaro et al. (2020) recommended a machine learning based method to predict the remaining useful life of military vehicles. This research demonstrated the potential of machine learning approaches in predicting equipment failures and optimizing maintenance schedules for armoured vehicles\cite{tessaro2020machine}.
Several studies have explored the potential of machine learning approaches for predictive maintenance in various industrial sectors. For example, a study by Paolanti et al. (2018) used a Random Forest-based Machine Learning architecture for Predictive Maintenance. The training data was collected form various sensors, machine PLCs and other components\cite{paolanti2018machine}. Similarly, a research by Divya et al. (2022) outlined a ML-based system for wind turbine predictive maintenance that achieved remarkably high accuracy. It highlighted machine learning's promise for predictive maintenance and highlighted its capacity to identify and stop equipment faults.\cite{divya2022review}.
Souza et al. (2022) conducted a review of machine learning techniques for predictive maintenance of industrial equipment, including armoured vehicles. They concluded that machine learning approaches, such as decision trees, random forests, and support vector machines, are effective in predicting equipment failures and can be used to improve the maintenance practices of armoured vehicles\cite{souza2021deep}.
Zhang et al. (2019) conducted a review of predictive maintenance techniques in industrial applications, including the use of machine learning approaches. They focused on data-driven methods for PdM. They presented six machine learning and deep learning (DL) algorithms are used to categorise specific industrial applications, and five performance metrics were compared for each classification\cite{zhang2019data}.
Raja et al. (2022) discussed the predictive maintenance(PdM) of various electrical machines, such as BLDC motors. To have a cost effective diagnostic system they presented a data acquisition system used to transmit the data in real-time onto the cloud, where it is further processed to ascertain whether there is a possibility that a motor fault could occur. They used IoT and Cloud techniques to maintain the different electrical components\cite{raja2022cost}.

\section{METHODOLOGY}
The methodology for this work consists of three main steps. Firstly, a dataset is extracted from the armoured vehicles using sensors and other monitoring equipment. We have taken the AI4I 2020 Predictive Maintenance Dataset for our study. Secondly, the data set is pre-processed to remove any noisy or irrelevant data, and class imbalance is addressed. Finally, the proposed ensemble model, consisting of multiple machine learning algorithms, is trained on the pre-processed dataset to predict potential equipment failures. The efficiency of the ensemble model in foreseeing maintenance difficulties is assessed using key measures.

\subsection{Dataset}\label{AA}
The AI4I 2020 Predictive Maintenance Dataset is a publicly available dataset provided by the UCI Machine Learning Repository\cite{torcianti2021explainable}. The dataset contains sensor data collected from an industrial production line of a simulated manufacturing plant. The purpose of the dataset is to facilitate research and development in predictive maintenance and machine learning. \\
The dataset comprises 10,000 rows of data, with six features stored in columns.

\begin{itemize}
    \item The first feature is the \emph{Product ID/Type}.
    \item The second feature is \emph{Air Temperature}, which is generated using a random walk process.
    \item The third feature is \emph{Process Temperature}.
    \item The fourth feature is \emph{Rotational Speed}. This is derived from a power of 2900 Watts.
    \item The fifth feature is \emph{Torque}, with values generally distributed about 40 Nm.
    \item The final feature is \emph{Tool Wear}, which is influenced by the product quality variant.
\end{itemize}
This synthetic dataset can be used to train and test machine learning models for predictive maintenance analysis, despite being a simulated representation of real-world maintenance data.

\subsection{Machine Failure Modes and Description}
 There are five distinct failure modes in the AI4I 2020 Predictive Maintenance Dataset\cite{torcianti2021explainable}. Each of the modes are listed below:
\begin{itemize}
    \item Tool Wear Failure (TWF)
    \item Heat Dissipation Failure (HDF)
    \item Power Failure (PWF)
    \item Overstrain Failure (OSF)
    \item Random Failures (RNF)
\end{itemize}
The machine failure label is assigned to 1 if any of the aforementioned procedures fail, which is the case for 339 data points. If none of the aforementioned failure types are present, the Machine Failure Label is set to 0.\cite{torcianti2021explainable}.\\
However, in this research only six features have been considered and a binary classification model is trained on them. A multi-label classification problem can be solved by including above five independent failure modes. But, it is not a part of the study.

\subsection{Feature Extraction}
As stated earlier, a total of six features were extracted from the dataset and have been displayed in Table \ref{tab1} with the respective values of all 10000 machines and their respective units. These features have been abbreviated to the following names: \emph{Type} as F1, \emph{Air Temperature} as F2, \emph{Process Temperature} as F3, \emph{Rotational Speed} as F4, \emph{Torque} as F5 and \emph{Tool Wear} as F6. 
\subsection{Data Preprocessing}
\label{preprocess}
In Statistical Machine Learning, the data must be pre-processed before training the machine learning model on it. The following steps were performed to preprocess the data:
\begin{itemize}
    \item The data along with the features and the output category was first represented in the form of a Pandas dataframe.
    \item The dataset was then divided into feature vector and output vector. Min-Max Normalization\cite{patro2015normalization} was used to scale the features F2, F3, F4, F5 and F6 (as shown in Table \ref{tab1}). The formula for this Normalization is shown in Equation \ref{eq1}, with $F_\alpha$ referring to the $\alpha$th feature.
    \begin{equation} \label{eq1}
    \begin{split}
    Scaled Data=(X[:,F_\alpha]- 
    min(X[:,F_\alpha]))\\
    /(max(X[:,F_\alpha])-min(X[:,F_\alpha]))
    \end{split}
    \end{equation}
    \item Once the Normalization was done, F1 was label encoded into three categories as shown in Equation \ref{eq2}
    \begin{equation} \label{eq2}
    X['Type']=[L,M,H] \rightarrow \\
    X['Type']=[1,2,0]\\
    \end{equation}
    \item The dataframe was then converted into 2-D NumPy array and was divided into a 70:30 train-test ratio.
    \item After balancing the data (described below), various models depicted in table \ref{tab2} were trained using Scikit-learn library and their classification scores were recorded.
\end{itemize}
\subsection{Data Balancing}
After splitting the dataset (70:30), there were only 240 instances of Label '1' as compared to the 6760 samples of Label '0'. Therefore, Synthetic Minority Over-sampling Technique (SMOTE) was used to equalize the number of both the samples.\\
\begin{figure}[ht]
\centerline{\includegraphics[scale=0.55]{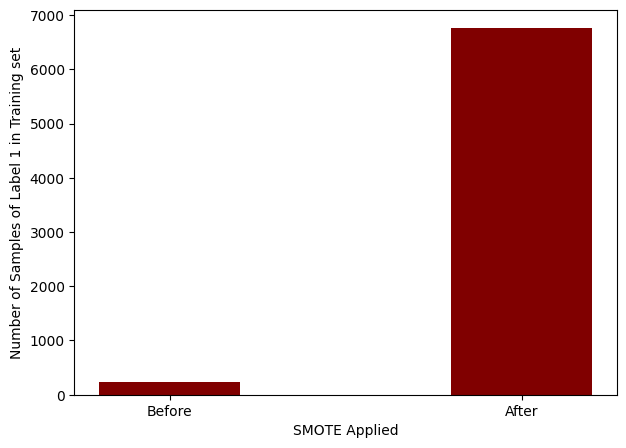}}
\caption{Variation in the count of minority class}
\label{fig_l1}
\end{figure}
By generating synthetic samples for the minority class, it helps in creating a more balanced dataset, leading to more robust and unbiased models. Moreover, SMOTE does not result in any loss of information since the synthetic instances are created from existing data points.
  \begin{equation} \label{eq_new}
   {x}_{new} = {x}_{min} + ({x}_i - {x}_{min}) \times \delta
    \end{equation}
In equation \ref{eq_new}, ${x}_{min}$ represents the number of original minority class datapoints. $ith$ nearest neighbor is depicted by ${x}_{i}$ and ${x}_{new}$ shows the synthetic datapoint generated.\\
As shown in figure \ref{fig_l1}, the count of label '1' is increased from 240 to 6760. Consequently, there is an increase of total samples in training set from 7000 to 13520 as shown by figure \ref{fig_tot}.
\begin{figure}[ht]
\centerline{\includegraphics[scale=0.55]{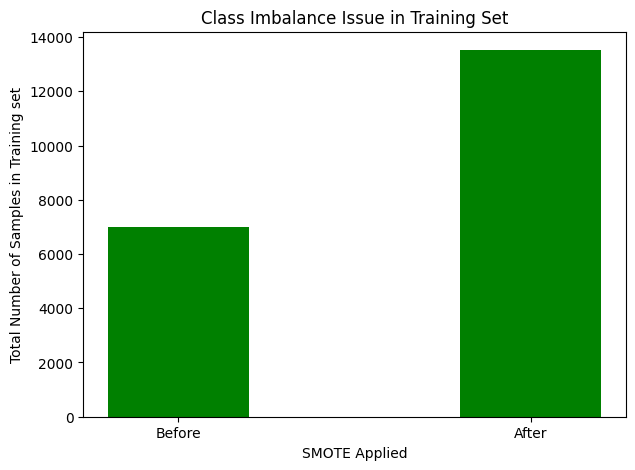}}
\caption{Variation in Training set (Total) }
\label{fig_tot}
\end{figure}

\begin{figure*}
\centerline{\includegraphics[scale=0.65]{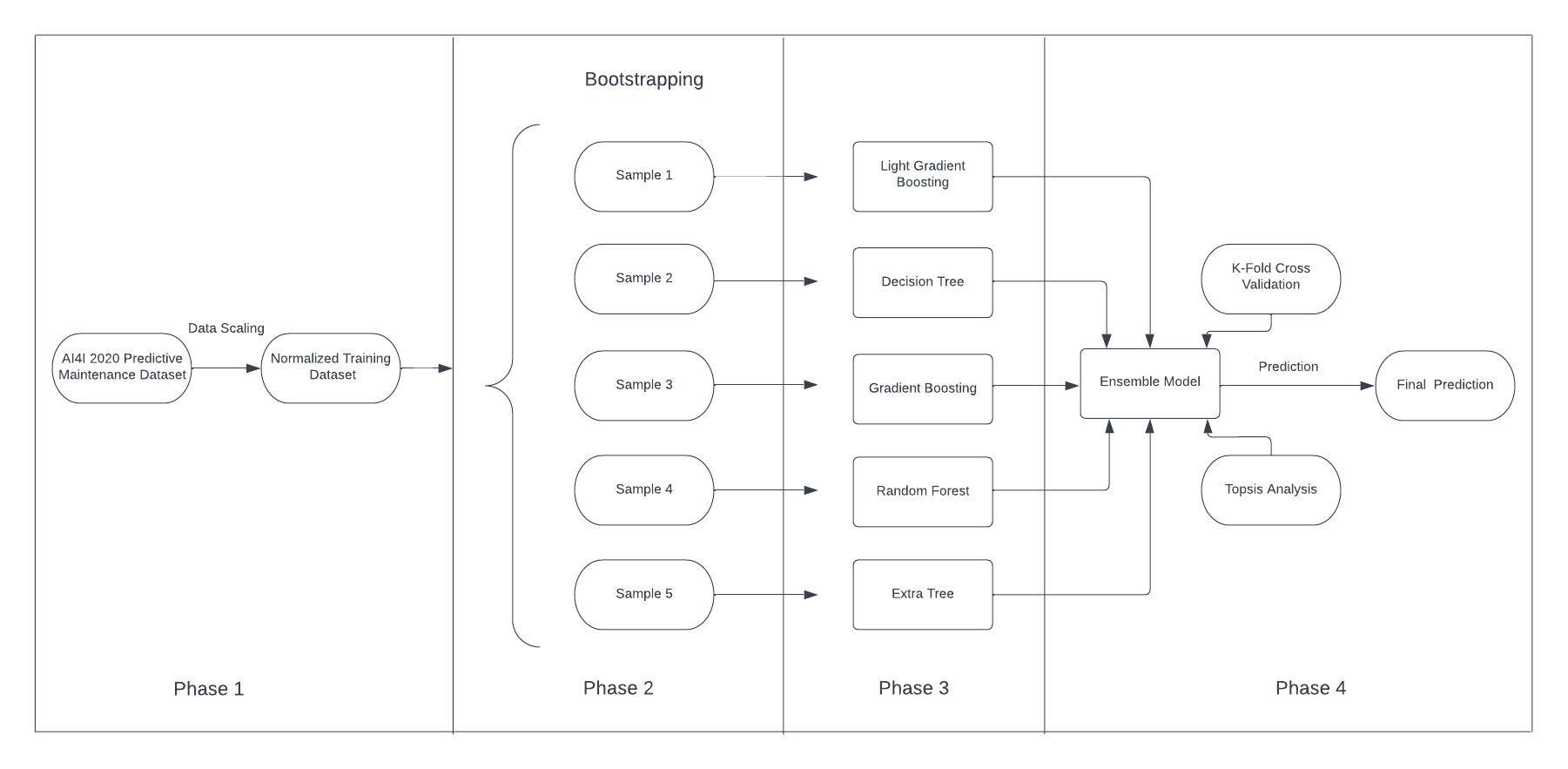}}
\caption{Workflow of the Proposed Ensemble Model}
\label{fig2}
\end{figure*}

\section{Proposed Model}
The pipeline's workflow and the working of the proposed ensemble model can be divided into four phases, as illustrated in Fig.\ref{fig2}. \\

\textbf{Phase 1}: In the first phase, we extracted the data from the AI4I 2020  Predictive Maintenance Dataset\cite{torcianti2021explainable}. Six features (F1, F2, F3, F4, F5 and F6) as depicted in table \ref{tab1} were extracted from the dataset. The data was than pre-processed as explained in Section \ref{preprocess}. The training and testing data was split in 70:30 ratio and datapoints were balanced.\\

\textbf{Phase 2}: Further, five bootstrap samples were created with replacement. The size of each bootstrap sample was kept as the size of the training data. These samples were to be used as input to five classifiers. These classification algorithms were decided in the next phase.\\

\textbf{Phase 3}: According to the accuracy obtained after hyper parameter tuning of the various Machine Learning Models shown in table \ref{tab2}, five best machine learning models were selected namely, Light Gradient Boosting, Decision Tree Classifier, Gradient Boosting Classifier, Random Forest Classifier and Extra Tree Classifier. Accordingly, five Bootstrap samples from the pre-processed were taken in this phase and fed onto the respective models.
The prediction was done according to majority voting.\\

\begin{table*}[ht]
\centering
\caption{Machine Learning Classification Models}
\begin{tabular}{ |p{1cm}|p{5cm}|p{5cm}|p{5cm}| }
\hline
\textbf{SN}& \textbf{Model}                           &  \textbf{Hyperparameters }                            \\ \hline
1.&Light Gradient Boosting Machine & max\_depth:20,number of iterations:50      \\ \hline
2.&Gradient Boosting Classifier    & min\_samples\_split:4,min\_samples\_leaf:1 \\ \hline
3.&Random Forest Classifier        & n\_estimators:100,criterion:'gini'         \\ \hline
4.&Decision Tree Classifier        & min\_samples\_split:2                      \\ \hline
5.&Extra Trees Classifier          & n\_estimators:100                          \\ \hline
6.&K Neighbors Classifier          & metric:'minkowski',leaf\_size:20           \\ \hline
7.&Ada Boost Classifier            & learning\_rate:0.5,n\_estimators:50        \\ \hline
8.&Linear Discriminant Analysis    & shrinkage:'auto'                           \\ \hline
9.&Logistic Regression             & penalty:'l2',tol:0.0001                    \\ \hline
10.&SVM - Linear Kernel             & kernel:'rbf,gamma:'scale'                  \\ \hline
11.&Ridge Classifier                & alpha:1                                    \\ \hline
12.&Dummy Classifier                & strategy:'most\_frequent'                  \\ \hline
13.&Naïve Bayes            & alpha:1                                    \\ \hline
14.&Quadratic Discriminant Analysis & tol:0.0001                                 \\ \hline
\end{tabular}
\label{tab2}
\end{table*}
\textbf{Phase 4}: Topsis Analysis was used to create the ranking of different models along with the ensemble model. The proposed ensemble achieved the first rank in the statistical analysis. The model's reliability was tested with additional K-Fold Cross Validation. (discussed in Section \ref{crossval}).

\begin{figure}[ht]
\centerline{\includegraphics[scale=0.6]{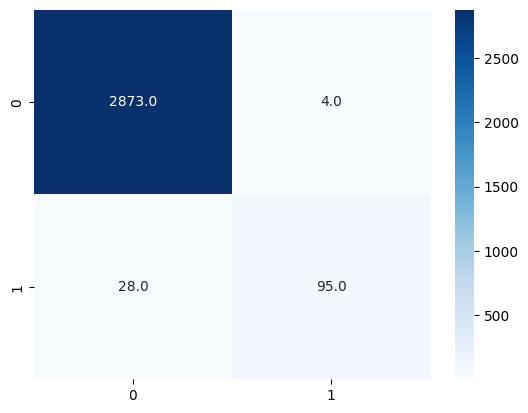}}
\caption{Confusion Matrix (Testing Set)}
\label{fig1}
\end{figure}

\section{Model Evaluation}
To assess the effectiveness of the suggested ensemble model, several parameters, including precision, recall, accuracy, AUC and F1, were calculated. The obtained metrics are organised in a table and are displayed in Table \ref{tab3}.

\begin{table*}
\centering
\caption{Evaluation Parameters}
\begin{tabular}{|c|c|c|c|c|c|c|}
\hline
\textbf{SN}  & \textbf{Model}                            & \textbf{Accuracy}  & \textbf{AUC}     & \textbf{Recall}  & \textbf{Precision} & \textbf{F1}    \\ \hline
1.     & Light Gradient Boosting Machine & 0.9847   & 0.9768 & 0.6215 & 0.8748    & 0.7208    \\ \hline
2.           & Decision Tree Classifier        & 0.981    & 0.8539 & 0.718  & 0.7103    & 0.7095   \\ \hline
3.           & Extra Trees Classifier          & 0.9777   & 0.9602 & 0.3344 & 0.9456    & 0.4906  \\ \hline
4.          & K Neighbors Classifier          & 0.9741   & 0.8194 & 0.268  & 0.8194    & 0.3952   \\ \hline
5.          & Ada Boost Classifier            & 0.9727   & 0.9548 & 0.401  & 0.6216    & 0.4834    \\ \hline
6.          & Quadratic Discriminant Analysis & 0.8319   & 0.807  & 0.4547 & 0.2939    & 0.2889    \\ \hline
7.          & Gradient Boosting Classifier    & 0.9821   & 0.9734 & 0.5911 & 0.8102    & 0.6796   \\ \hline
8.           & Random Forest Classifier        & 0.9817   & 0.9654 & 0.4978 & 0.9018    & 0.6351   \\ \hline
9.         & Linear Discriminant Analysis    & 0.9687   & 0.8748 & 0.3573 & 0.5413    & 0.4241    \\ \hline
10.           & Logistic Regression             & 0.968    & 0.8472 & 0.0132 & 0.2       & 0.0247   \\ \hline
11.        & Dummy Classifier                & 0.9676   & 0.5    & 0      & 0         & 0         \\ \hline
12.          & Naïve Bayes                     & 0.961    & 0.8696 & 0.2241 & 0.3451    & 0.27      \\ \hline
\textbf{13.}         & \textbf{Proposed Ensemble Model} & \textbf{0.9893}    & \textbf{0.957}  & \textbf{0.9903} &  \textbf{0.9980}   & \textbf{0.9941}
\\ \hline

\end{tabular}
\label{tab3}
\end{table*}

\subsection{Model Evaluation Parameters}
\label{modeleval}
\begin{enumerate}
  \item \emph{Precision}: Precision measures how often the model's positive predictions are correct. It is calculated by dividing the entire number of the true positives (TP) by the addition of the true positives and false positives (FP). Precision is computed as:
    \begin{equation} \label{eq3}
    Precision = \gamma / (\gamma + \lambda)
    \end{equation}
    where $\gamma$ is the number of accurate results and $\lambda$ is the number of erroneous ones.
 \item \emph{F1 Score}: The F1 score is a measurement that combines recall and precision into a single figure. It is determined by averaging them harmonically and is calculated using the following equation:
    \begin{equation} \label{eq4}
    F1 = 2 * (a * b) / (a + b)
    \end{equation}  
wherein \emph{a} corresponds to Precision and \emph{b }corresponds to the Recall obtained by training the classification model.
 \item \emph{Area under curve (AUC)}: AUC is a measure of the model's ability to distinguish between positive and negative classes. AUC measures the model's ability to distinguish between positive and negative cases. The area under curve constructed by plotting the true positive rate (TPR) against the false positive rate (FPR) at varying cutoff values is the measure of accuracy. 
  \item \emph{Recall}: Recall measures how well the model is able to identify all the positive cases. It is determined by dividing the total number of positive results ($\gamma$) by the total number of possible negative results ($\gamma$+$\chi$). The following equation is used to determine it:
    \begin{equation} \label{eq5}
    Recall = \gamma / (\gamma + \chi)
    \end{equation} 
   \item \emph{Accuracy}: The accuracy of a model is evaluated by how well its predictions actually turn out. Correct predictions are divided by overall predictions to get this ratio. It is calculated using the following formula:
    \begin{equation} \label{eq7}
        Accuracy = (TP + TN) / (TP + TN + FP + FN)
    \end{equation}
 Correct predictions are the addition of TP and TN, overall resulting prediction is the collective summation of TP, TN, FP and FN. 
\subsection{Topsis}

\begin{table}
\centering
\caption{Topsis}
\begin{tabular}{|l|l|l|l|}
\hline
\textbf{SN} & \textbf{Model}                           & \textbf{Rank} &  \textbf{Score} \\
\hline
1.    & Proposed Ensemble Model         & 1    & 0.96  \\
2.    & Light Gradient Boosting Machine & 2    & 0.89  \\
3.    & Gradient Boosting Classifier    & 3    & 0.89  \\
4.    & Random Forest Classifier        & 4    & 0.87  \\
5.    & Decision Tree Classifier        & 5    & 0.87  \\
6.    & Extra Trees Classifier          & 6    & 0.85  \\
7.    & K Neighbors Classifiers         & 7    & 0.84  \\
8.    & Ada Boost Classifier            & 8    & 0.83  \\
9.    & Linear Discriminant Analysis    & 9    & 0.82  \\
10.   & Logistic Regression             & 10   & 0.82  \\
11.   & Dummy Classifier                & 11   & 0.79  \\
12.   & Naïve Bayes                     & 12   & 0.71  \\
13.   & Quadratic Discriminant Analysis & 13   & 0.6  \\
\hline

\end{tabular}
\end{table}
TOPSIS is a multi-criteria decision analysis method used for solving complex decision-making problems. It is a commonly used tool in operations research and management sciences.

The basic idea behind TOPSIS is to identify the best alternative out of a set of alternatives based on a set of criteria. The method evaluates each alternative based on how well it satisfies the criteria and then ranks the alternatives in order of preference.

Each TOPSIS criteria contains both a positive and a negative ideal solution that are necessary for the approach to work. When evaluating a set of criteria, the best possible answer is the ideal solution, while the worst possible solution is the negative ideal solution. The ideal and negative ideal solutions are determined by the decision maker based on their preferences and goals\cite{behzadian2012state}.

\subsection{K-Fold Cross Validation}
\label{crossval}
A common method for assessing a machine learning model's performance is K-fold cross-validation. It involves dividing the data into k equal-sized subsets, or folds. Out of the k folds, each fold is once used for testing and the remaining k-1 folds used for training. Repeated K-fold Cross Validation is used to test whether the ensemble model that is being suggested is consistent with low bias and variance\cite{browne2000cross}. Five instances of the 5-fold Cross Evaluation are performed in the current study. The resulting graph of the above cross validation is shown in figure \ref{fig3}. As the lines are coinciding, it is indicating that the suggested ensemble model is reliable. The overall average accuracy after five trials is 95.36\%.

\begin{figure}
\centerline{\includegraphics[scale=0.6]{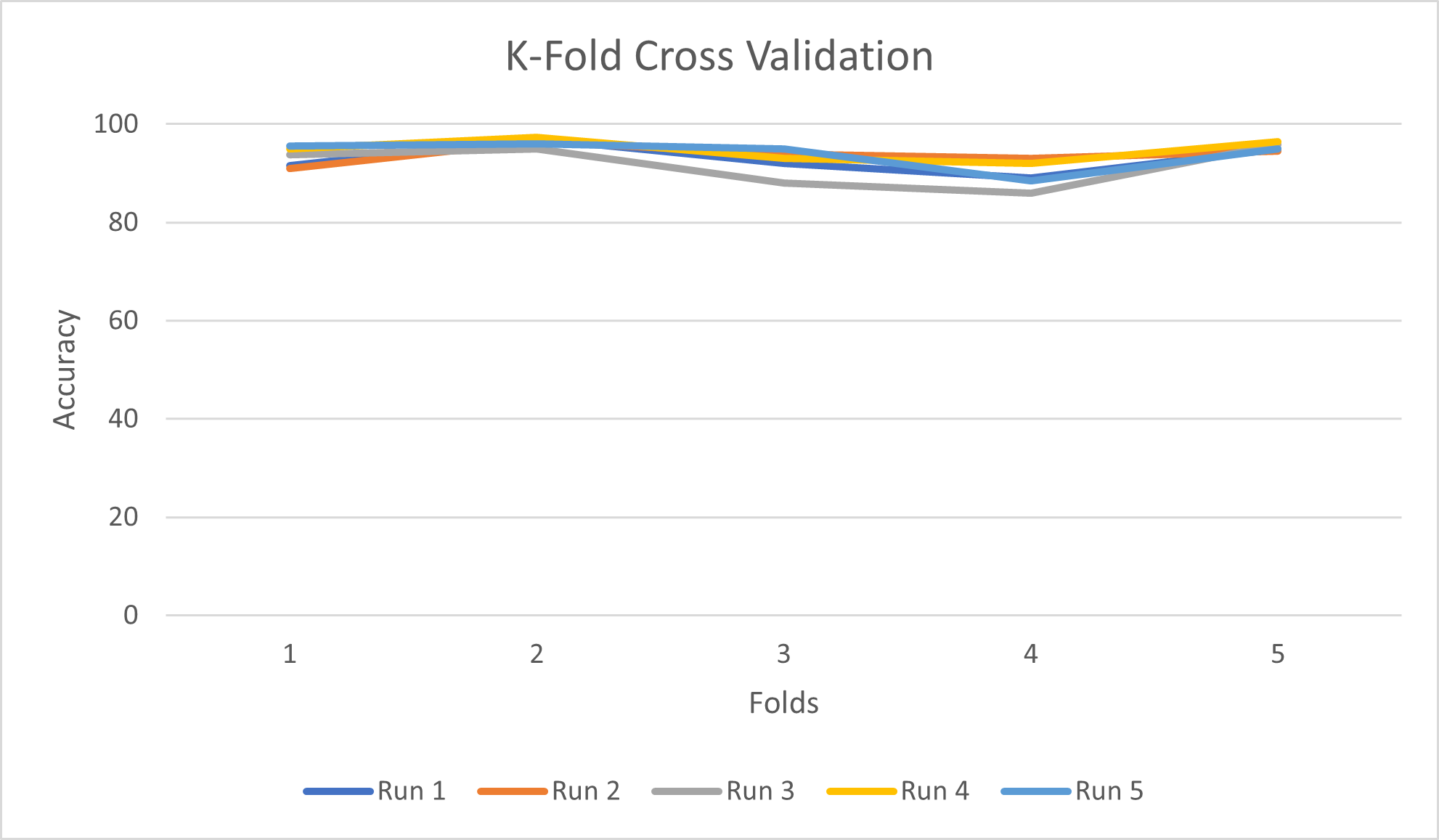}}
\caption{K-Fold Cross Validation}
\label{fig3}
\end{figure}
\end{enumerate}

\section{Result Analysis and Discussion}

The machine learning models that are shown in \ref{tab2} were trained using the AI4I 2020 Predictive Maintenance Dataset along with the hyper parameters that were adjusted for each model accordingly.
The dataset for training was used to train the models by taking the bootstrap sample (SWR), and the testing dataset was used to validate them. Five models were combined in the ensemble model that is being proposed. The models were evaluated according to the evaluation parameter mentioned in Section \ref{modeleval}. The proposed ensemble model outperformed other machine learning models, according to Topsis Analysis as well as in the evaluation parametric.\\ 
The proposed algorithm achieves a testing accuracy of 98.93\%, precision of 99.80\% and a recall of 99.03\%.
Overfitting is a potential issue that can arise during training. To overcome this we had cross-validated the model using K-Fold cross-validations with 5 runs.

\section{Conclusion and Future Scope}
In conclusion, predictive maintenance using machine learning approaches has the potential to significantly improve the reliability and availability of armoured vehicles. By analyzing large amounts of data, such as sensor readings, historical maintenance records, and operational conditions, machine learning models can identify patterns and anomalies that indicate potential failures or maintenance needs before they occur.
Moreover, the application of machine learning based approaches can help reduce costs associated with unexpected breakdowns, extend the lifespan of armoured vehicles, and improve overall mission readiness. The proposed ensemble model created using Decision Tree, Random Forest, Extra Tree Classifier and Gradient Boosting techniques achieves an accuracy of 98.93\%. Future research in this field may concentrate on enhancing the predictive maintenance models' accuracy by including further data sources, such as weather information, information about the topography, and operator behavior. With ongoing research and development efforts aimed at improving the accuracy and efficiency of machine learning algorithms. The use of real-time data from sensors and the incorporation of advanced analytics, such as deep learning, are expected to enhance the ability of predictive maintenance systems to detect and anticipate problems. Additionally, research could be conducted to develop new machine learning algorithms that can operate with limited or incomplete data, as well as explore ways to integrate these models into existing maintenance management systems. Also, finer tuning of the hyperparameters as displayed in Table \ref{tab2} may be done to raise the suggested ensemble model's accuracy. 

\bibliographystyle{unsrt}
\bibliography{bibfile}

\begin{thebibliography}{10}

\bibitem{borkowski2010influence}
Wac{\l}aw Borkowski, Piotr Rybak, Zdzis{\l}aw Hryci{\'o}w, J{\'o}zef Wysocki,
  and Bogus{\l}aw Micha{\l}owski.
\newblock Influence of operation conditions on the wheeled armoured carrier
  characteristics.
\newblock {\em Journal of KONES}, 17:59--65, 2010.

\bibitem{dong2020survey}
Xibin Dong, Zhiwen Yu, Wenming Cao, Yifan Shi, and Qianli Ma.
\newblock A survey on ensemble learning.
\newblock {\em Frontiers of Computer Science}, 14:241--258, 2020.

\bibitem{xiao2021review}
Zhuo Xiao, Zhe Cheng, and Yuehao Li.
\newblock A review of fault diagnosis methods based on machine learning
  patterns.
\newblock {\em 2021 Global Reliability and Prognostics and Health Management
  (PHM-Nanjing)}, pages 1--4, 2021.

\bibitem{theissler2021predictive}
Andreas Theissler, Judith P{\'e}rez-Vel{\'a}zquez, Marcel Kettelgerdes, and
  Gordon Elger.
\newblock Predictive maintenance enabled by machine learning: Use cases and
  challenges in the automotive industry.
\newblock {\em Reliability engineering \& system safety}, 215:107864, 2021.

\bibitem{arena2022predictive}
Fabio Arena, Mario Collotta, Liliana Luca, Marianna Ruggieri, and
  Francesco~Gaetano Termine.
\newblock Predictive maintenance in the automotive sector: A literature review.
\newblock {\em Mathematical and Computational Applications}, 27(1):2, 2022.

\bibitem{putra2021designing}
Hafid Galih~Pratama Putra, Suhono~Harso Supangkat, I~Gusti Bagus~Baskara
  Nugraha, Fadhil Hidayat, and PT~Kereta.
\newblock Designing machine learning model for predictive maintenance of
  railway vehicle.
\newblock In {\em 2021 International Conference on ICT for Smart Society
  (ICISS)}, pages 1--5. IEEE, 2021.

\bibitem{jain2022systematic}
Muskan Jain, Dipit Vasdev, Kunal Pal, and Vishal Sharma.
\newblock Systematic literature review on predictive maintenance of vehicles
  and diagnosis of vehicle's health using machine learning techniques.
\newblock {\em Computational Intelligence}, 38(6):1990--2008, 2022.

\bibitem{tessaro2020machine}
Iron Tessaro, Viviana~Cocco Mariani, and Leandro dos~Santos Coelho.
\newblock Machine learning models applied to predictive maintenance in
  automotive engine components.
\newblock In {\em Proceedings}, volume~64, page~26. MDPI, 2020.

\bibitem{paolanti2018machine}
Marina Paolanti, Luca Romeo, Andrea Felicetti, Adriano Mancini, Emanuele
  Frontoni, and Jelena Loncarski.
\newblock Machine learning approach for predictive maintenance in industry 4.0.
\newblock In {\em 2018 14th IEEE/ASME International Conference on Mechatronic
  and Embedded Systems and Applications (MESA)}, pages 1--6. IEEE, 2018.

\bibitem{divya2022review}
D~Divya, Bhasi Marath, and MB~Santosh~Kumar.
\newblock Review of fault detection techniques for predictive maintenance.
\newblock {\em Journal of Quality in Maintenance Engineering}, 2022.

\bibitem{souza2021deep}
Roberto~M Souza, Erick~GS Nascimento, Ubatan~A Miranda, Wenisten~JD Silva, and
  Herman~A Lepikson.
\newblock Deep learning for diagnosis and classification of faults in
  industrial rotating machinery.
\newblock {\em Computers \& Industrial Engineering}, 153:107060, 2021.

\bibitem{zhang2019data}
Weiting Zhang, Dong Yang, and Hongchao Wang.
\newblock Data-driven methods for predictive maintenance of industrial
  equipment: A survey.
\newblock {\em IEEE Systems Journal}, 13(3):2213--2227, 2019.

\bibitem{raja2022cost}
Hadi~Ashraf Raja, Hardik Raval, Toomas Vaimann, Ants Kallaste, Anton
  Rass{\~o}lkin, and Anouar Belahcen.
\newblock Cost-efficient real-time condition monitoring and fault diagnostics
  system for bldc motor using iot and machine learning.
\newblock In {\em 2022 International Conference on Diagnostics in Electrical
  Engineering (Diagnostika)}, pages 1--4. IEEE, 2022.

\bibitem{torcianti2021explainable}
Andrea Torcianti and Stephan Matzka.
\newblock Explainable artificial intelligence for predictive maintenance
  applications using a local surrogate model.
\newblock In {\em 2021 4th International Conference on Artificial Intelligence
  for Industries (AI4I)}, pages 86--88. IEEE, 2021.

\bibitem{patro2015normalization}
SGOPAL Patro and Kishore~Kumar Sahu.
\newblock Normalization: A preprocessing stage.
\newblock {\em arXiv preprint arXiv:1503.06462}, 2015.

\bibitem{behzadian2012state}
Majid Behzadian, S~Khanmohammadi Otaghsara, Morteza Yazdani, and Joshua
  Ignatius.
\newblock A state-of the-art survey of topsis applications.
\newblock {\em Expert Systems with applications}, 39(17):13051--13069, 2012.

\bibitem{browne2000cross}
Michael~W Browne.
\newblock Cross-validation methods.
\newblock {\em Journal of mathematical psychology}, 44(1):108--132, 2000.

\end{thebibliography}
\vspace{12pt}

\end{document}